\documentclass{llncs}
\usepackage[sort,square,comma,sectionbib]{natbib}
\usepackage{mathabx}
\usepackage{ifpdf}
\ifpdf
 \usepackage[pdftex]{graphicx}
\else
 \usepackage[dvips]{graphicx}
\fi
\usepackage{tikz}
\usepackage{url}
\usepackage{xcolor}

\begin{document}
\title{Towards Understanding Triangle Construction Problems\thanks{This work was partially
supported by the Serbian Ministry of Science grant 174021 and by Swiss National Science
Foundation grant SCOPES IZ73Z0\_127979/1.}}
\author{Vesna Marinkovi\'c \and Predrag Jani\v{c}i\'c}
\institute{Faculty of Mathematics, University of Belgrade, Serbia
}

\maketitle

\begin{abstract}
Straightedge and compass construction problems are one of the oldest and
most challenging problems in elementary mathematics. The central challenge,
for a human or for a computer program, in solving construction problems
is a huge search space. In this paper we analyze one family of triangle
construction problems, aiming at detecting a small core of the
underlying geometry knowledge. The analysis leads to a small set of
needed definitions, lemmas and primitive construction steps, and
consequently, to a simple algorithm for automated solving of problems
from this family. The same approach can be applied to other families
of construction problems.
\end{abstract}

\begin{keywords}
Triangle construction problems, automated deduction in geometry
\end{keywords}

\section{Introduction}
\label{sec:introduction}

\begin{figure}
\begin{center}
\input{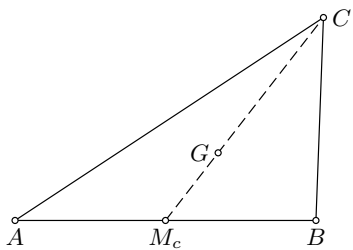}
\end{center}
\vspace*{-6mm}
\caption{Construction of a triangle $ABC$ given its vertices $A$, $B$ and
the barycenter $G$}
\label{fig:barycenter}
\end{figure}

Triangle construction problems (in Euclidean plane) are problems in which
one has to construct, using straightedge\footnote{The notion of ``straightedge''
is weaker than ``ruler'', as ruler is assumed to have markings which could be used
to make measurements. Geometry constructions typically require use of
straightedge and compass, not of ruler and compass.}
and compass,\footnote{By compass, we mean {\em collapsible compass}.
In contrast to {\em rigid compass}, one cannot use collapsible compass to
``hold'' the length while moving one point of the compass to another point.
One can only use it to hold the radius while one point of the compass is
fixed \cite{Beeson2010}.} a triangle that meets given (usually three)
constraints \cite{Martin,Lopes,constructions-teamat}.
The central problem, for a human or for a computer program, in solving triangle
construction problems is a huge search space: primitive construction steps
can be applied in a number of ways, exploding further along the construction.
Consider, as an illustration, the following simple problem:
{\em given the points $A$, $B$, and $G$, construct a triangle $ABC$ such that
$G$ is its barycenter}. One possible solution is: construct the midpoint $M_c$
of the segment $AB$ and then construct a point $C$ such that
$\widearrow{M_cG}/\widearrow{M_cC}=1/3$
(Figure \ref{fig:barycenter}). The solution is very simple and intuitive. However, if one
wants to describe a systematic (e.g., automatic) way for reaching this solution,
he/she should consider a wide range of possibilities. For instance, after
constructing the point $M_c$, one might consider constructing midpoints of the
segments $AG$, $BG$, or even of the segments $AM_c$, $BM_c$, $GM_c$, then midpoints
of segments with endpoints among these points, etc. Instead of the step that introduces
the point $C$ such that $\widearrow{M_cG}/\widearrow{M_cC}=1/3$ one may
(unnecessarily) consider introducing a point $X$ such that $\widearrow{AG}/\widearrow{AX}=1/3$ or
$\widearrow{BG}/\widearrow{BX}=1/3$. Also, instead of the step that introduces
the point $C$ such that $\widearrow{M_cG}/\widearrow{M_cC}=1/3$ one may
consider introducing a point $X$ such that $\widearrow{M_cG}/\widearrow{M_cX}=k$,
where $k \neq 1/3$, etc. Therefore, this trivial example shows that any systematic
solving of construction problems can face a huge search space even if only two
high-level constructions steps that are really needed are considered.
Additional problem in solving construction problems makes the fact that some of
them are unsolvable (which can be proved by an algebraic argument), including, for
instance, three antiquity geometric problems: circle squaring, cube duplication,
angle trisection \cite{GaloisTheory}. Although the problem of constructibility (using
straightedge and compass) of a figure that can be specified by algebraic equations
with coefficients in ${\bf Q}$ is decidable \cite{Lebesgue,GaoC98a,Guoting}, there
are no simple and efficient, ``one-button'' implemented decision procedures so,
typically, proofs of insolvability of construction problems are made {\em ad-hoc}
and not derived by uniform algorithms.
\vspace*{-7mm}

\vspace*{-4mm}

Construction problems have been studied, since ancient Greeks, for centuries and
represent a challenging area even for experienced mathematicians. Since early
twentieth century, ``triangle geometry'', including triangle construction problems,
has not been considered a premier research field \cite{DavisTriangleGeometry}.
However, construction problems kept their role on all levels of mathematical
education, almost in the same form for more than two and a half millenia,
which make them probably the problems used most constantly throughout the
history of mathematical education. Since the late twentieth century, geometry
constructions are again a subject of research, but this time mainly meta-mathematical.
There are two main lines of this work:
\begin{itemize}
\item {\bf Research in axiomatic foundations of geometry constructions and foundational
issues.} According to Pambuccian and his survey of axiomatizing geometric constructions,
surprisingly, it is only in 1968 that geometric constructions became part of
axiomatizations of geometry \cite{Pambuccian08}.
In constructive geometry axiomatizations, following ancient understanding, axioms
are quantifier-free and involve only operation symbols (reflecting construction steps)
and individual constants, in contrast to the Hilbert-style approach with relation symbols
and where axioms are not universal statements. One such axiomatic theory of constructive
geometry --- {\bf ECG} (``Elementary Constructive Geometry'') was recently proposed by
Beeson \cite{Beeson2010}. Constructive axiomatizations bring an alternative approach in
geometry foundations, but they do not bring a substantial advantage to the Hilbert
style when it comes to solving concrete construction problems.

\item {\bf Research in developing algorithms for solving construction problems.}
There are several approaches for automated solving of construction problems
\cite{GaoC98,SchreckPhD,GulwaniKT11,Grima}. However, most, if not all of
them, focus on search procedures and do not focus on finding a small portion
of geometry knowledge (axioms and lemmas) that are underlying the constructions
(although, naturally, all approaches have strict lists of available primitive
construction steps). Earlier attempts at (manual) systematization of triangle
construction problems also didn't provide small and clear, possibly minimal in
some sense, lists of needed underlying geometry knowledge \cite{Lopes,Fursenko1,Fursenko2}.
\end{itemize}

We find that it is important to locate, understand and systematize
the knowledge relevant for solving construction problems or their subclasses.
That should be useful for teachers, students and mathematical knowledge base
generally. Also, such understanding should lead to a system that automatically
solves these kinds of problems (and should be useful in education).

In this work we focus on one family of triangle construction problems
and try to derive an algorithm for automated solving of problems based
on a small portion of underlying geometry knowledge. Our analyses lead us to a
small set of definitions, lemmas and construction rules needed for solving most
of the solvable problems of this family. The same approach can be applied to
other sorts of triangle construction problems and, more generally, to other
sorts of construction problems. The approach, leading to a compact representation
of the underlying geometry knowledge, can be seen not only as an algorithm
for automated solving of triangle construction problems but also as a work
in geometry knowledge management, providing a compact representation for a
large sets of construction problems, currently not available in the literature.

\section{Constructions by Straightedge and Compass}
\label{sec:constructions}

There are several closely related definitions of a notion of constructions
by straightedge and compass \cite{deTemple,GaloisTheory,Beeson2010}. By a
straightedge-and-compass construction we will mean a sequence of the
following primitive (or {\em elementary}) steps:
\begin{itemize}
\item construct an arbitrary point (possibly distinct from some given points);
\item construct (with {\em ruler}) the line passing through two given distinct points;
\item construct (with {\em compass}) the circle centered at some point passing through another point;
\item construct an intersection (if it exists) of two circles, two lines, or
a line and a circle.
\end{itemize}

In describing geometrical constructions, both primitive and compound constructions
can be used. A straightedge-and-compass construction problem is a problem in which
one has to provide a straightedge-and-compass construction such that the constructed
objects meet given conditions. A solution of a geometrical construction problem
traditionally includes the following four phases/components
\cite{Adler1906,Lopes,holland,constructions-teamat}:

\begin{description}
\item[Analysis:] In analysis one typically starts from the assumption
that a certain geometrical object satisfies the specification $\Gamma$
and proves that properties $\Lambda$ enabling the construction also hold.
\item[Construction:] In this phase, straightedge-and-compass
construction based on the analysis (i.e, on the properties $\Lambda$
which are proved within it) has to be provided.
\item[Proof:] It this phase, it has to be proved that the provided
straightedge-and-compass construction meets the given specification,
i.e., the conditions $\Gamma$.
\item[Discussion:]
In the discussion, it is considered how many possible solutions to
the problem there exist and under which conditions.
\end{description}

\section{Wernick's Problems}
\label{sec:wernickproblems}

In 1982, Wernick presented a list of triangle construction problems \cite{Wernick}.
In each problem, a task is to construct a triangle from three located points
selected from the following set of 16 characteristic points:
\begin{itemize}
\item $A$, $B$, $C$, $O$: three vertices and circumcenter;
\item $M_a$, $M_b$, $M_c$, $G$:  the side midpoints and barycenter (centroid);
\item $H_a$, $H_b$, $H_c$, $H$:  three feet of altitudes and orthocenter;
\item $T_a$, $T_b$, $T_c$, $I$:  three feet of the internal angles bisectors, and incenter;
\end{itemize}

There are 560 triples of the above points, but Wernick's list consists only of 139
significantly different non-trivial problems. The triple $(A,B,C)$ is trivial and,
for instance, the problems $(A, B, M_a)$, $(A, B, M_b)$, $(B, C, M_b)$, $(B, C, M_c)$,
$(A, C, M_a)$, and $(A, C, M_c)$ are considered to be symmetric (i.e., analogous). Some
triples are redundant (e.g., $(A,B,M_c)$ --- given points $A$ and $B$, the point
$M_c$ is uniquely determined, so it is redundant in $(A,B,M_c)$), but are still listed
and marked {\bf R} in Wernick's list. Some triples are constrained by specific conditions,
for instance, in $(A,B,O)$ the point $O$ has to belong to the perpendicular bisector of $AB$
(and in that case there are infinitely many solutions). In these problems, the locus restriction
gives either infinitely many or no solutions. These problems are marked {\bf L} in
Wernick's list. There are 113 problems that do not belong to the groups marked {\bf R}
and {\bf L}. Problems that can be solved by straightedge and ruler are marked {\bf S}
and problems that cannot be solved by straightedge and ruler are marked {\bf U}.

In the original list, the problem 102 was erroneously marked {\bf S} instead of
{\bf L} \cite{WernickUpdate}. Wernick's list left 41 problem unresolved/unclassified,
but the update from 1996 \cite{WernickUpdate} left only 20 of them.
In the meanwhile, the problems 90, 109, 110, 111 \cite{Specht-web}, and 138
\cite{Wernick138} were proved to be unsolvable. Some of the problems were
additionally considered for simpler solutions, like the problem 43
\cite{wernick43,wernick43a}, the problem 57 \cite{wernick57}, or the problem
58 \cite{wernick58,Specht-web}. Of course, many of the problems from Wernick's
list were considered and solved along the centuries without the context of this
list. The current status (to the best of our knowledge) of the problems from
Wernick's list is given in Table \ref{table:wernick-list}: there are 72 {\bf S}
problems, 16 {\bf U} problems, 3 {\bf R} problems, and 23 {\bf L} problems.
Solutions for 59 solvable problems can be found on the Internet \cite{Specht-web}.

\begin{table}[t!]
{\scriptsize
\begin{center}
\begin{tabular}{|llll||llll||llll||llll|} \hline
1.  &  $A$, $B$, $O$     & L&                      &  36. & $A$, $M_b$, $T_c$ & S&                     &  71. & $O$, $G$, $H$       & R&                      &    106.& $M_a$, $H_b$, $T_c$  & U & \cite{WernickUpdate} \\ \hline
2.  &  $A$, $B$, $M_a$   & S&                      &  37. & $A$, $M_b$, $I$   & S&                     &  72. & $O$, $G$, $T_a$     & U& \cite{WernickUpdate} &    107.& $M_a$, $H_b$, $I$    & U & \cite{WernickUpdate} \\ \hline
3.  &  $A$, $B$, $M_c$   & R&                      &  38. & $A$, $G$, $H_a$   & L&                     &  73. & $O$, $G$, $I$       & U& \cite{WernickUpdate} &    108.& $M_a$, $H$, $T_a$    & U & \cite{WernickUpdate} \\ \hline
4.  &  $A$, $B$, $G$     & S&                      &  39. & $A$, $G$, $H_b$   & S&                     &  74. & $O$, $H_a$, $H_b$   & U& \cite{WernickUpdate} &    109.& $M_a$, $H$, $T_b$    & U & \cite{Specht-web}    \\ \hline
5.  &  $A$, $B$, $H_a$   & L&                      &  40. & $A$, $G$, $H$     & S&                     &  75. & $O$, $H_a$, $H$     & S&                      &    110.& $M_a$, $H$, $I$      & U & \cite{Specht-web}    \\ \hline
6.  &  $A$, $B$, $H_c$   & L&                      &  41. & $A$, $G$, $T_a$   & S&                     &  76. & $O$, $H_a$, $T_a$   & S&                      &    111.& $M_a$, $T_a$, $T_b$  & U & \cite{Specht-web}    \\ \hline
7.  &  $A$, $B$, $H$     & S&                      &  42. & $A$, $G$, $T_b$   & U& \cite{WernickUpdate}&  77. & $O$, $H_a$, $T_b$   &  &                      &    112.& $M_a$, $T_a$, $I$    & S &                      \\ \hline
8.  &  $A$, $B$, $T_a$   & S&                      &  43. & $A$, $G$, $I$     & S& \cite{WernickUpdate}&  78. & $O$, $H_a$, $I$     &  &                      &    113.&  $M_a$, $T_b$, $T_c$ &   &                      \\ \hline
9.  &  $A$, $B$, $T_c$   & L&                      &  44. & $A$, $H_a$, $H_b$ & S&                     &  79. & $O$, $H$, $T_a$     & U& \cite{WernickUpdate} &    114.&  $M_a$, $T_b$, $I$   &  U& \cite{WernickUpdate} \\ \hline
10. &  $A$, $B$, $I$     & S&                      &  45. & $A$, $H_a$, $H$   & L&                     &  80. & $O$, $H$, $I$       & U& \cite{WernickUpdate} &    115.&  $G$, $H_a$, $H_b$   &  U& \cite{WernickUpdate} \\ \hline
11. &  $A$, $O$, $M_a$   & S&                      &  46. & $A$, $H_a$, $T_a$ & L&                     &  81. & $O$, $T_a$, $T_b$   &  &                      &    116.&  $G$, $H_a$, $H$     &  S&                      \\ \hline
12. &  $A$, $O$, $M_b$   & L&                      &  47. & $A$, $H_a$, $T_b$ & S&                     &  82. & $O$, $T_a$, $I$     & S& \cite{WernickUpdate} &    117.&  $G$, $H_a$, $T_a$   &  S&                      \\ \hline
13. &  $A$, $O$, $G$     & S&                      &  48. & $A$, $H_a$, $I$   & S&                     &  83. & $M_a$, $M_b$, $M_c$ & S&                      &    118.&  $G$, $H_a$, $T_b$   &   &                      \\ \hline
14. &  $A$, $O$, $H_a$   & S&                      &  49. & $A$, $H_b$, $H_c$ & S&                     &  84. & $M_a$, $M_b$, $G$   & S&                      &    119.&  $G$, $H_a$, $I$     &   &                      \\ \hline
15. &  $A$, $O$, $H_b$   & S&                      &  50. & $A$, $H_b$, $H$   & L&                     &  85. & $M_a$, $M_b$, $H_a$ & S&                      &    120.&  $G$, $H$, $T_a$     &  U& \cite{WernickUpdate} \\ \hline
16. &  $A$, $O$, $H$     & S&                      &  51. & $A$, $H_b$, $T_a$ & S&                     &  86. & $M_a$, $M_b$, $H_c$ & S&                      &    121.&  $G$, $H$, $I$       &  U& \cite{WernickUpdate} \\ \hline
17. &  $A$, $O$, $T_a$   & S&                      &  52. & $A$, $H_b$, $T_b$ & L&                     &  87. & $M_a$, $M_b$, $H$   & S& \cite{WernickUpdate} &    122.&  $G$, $T_a$, $T_b$   &   &                      \\ \hline
18. &  $A$, $O$, $T_b$   & S&                      &  53. & $A$, $H_b$, $T_c$ & S&                     &  88. & $M_a$, $M_b$, $T_a$ & U& \cite{WernickUpdate} &    123.&  $G$, $T_a$, $I$     &   &                      \\ \hline
19. &  $A$, $O$, $I$     & S&                      &  54. & $A$, $H_b$, $I$   & S&                     &  89. & $M_a$, $M_b$, $T_c$ & U& \cite{WernickUpdate} &    124.&  $H_a$, $H_b$, $H_c$ &  S&                      \\ \hline
20. &  $A$, $M_a$, $M_b$ & S&                      &  55. & $A$, $H$, $T_a$   & S&                     &  90. & $M_a$, $M_b$, $I$   & U& \cite{Specht-web}    &    125.&  $H_a$, $H_b$, $H$   &  S&                      \\ \hline
21. &  $A$, $M_a$, $G$   & R&                      &  56. & $A$, $H$, $T_b$   & U& \cite{WernickUpdate}&  91. & $M_a$, $G$, $H_a$   & L&                      &    126.&  $H_a$, $H_b$, $T_a$ &  S&                      \\ \hline
22. &  $A$, $M_a$, $H_a$ & L&                      &  57. & $A$, $H$, $I$     & S& \cite{WernickUpdate}&  92. & $M_a$, $G$, $H_b$   & S&                      &    127.&  $H_a$, $H_b$, $T_c$ &   &                      \\ \hline
23. &  $A$, $M_a$, $H_b$ & S&                      &  58. & $A$, $T_a$, $T_b$ & S& \cite{WernickUpdate}&  93. & $M_a$, $G$, $H$     & S&                      &    128.&  $H_a$, $H_b$, $I$   &   &                      \\ \hline
24. &  $A$, $M_a$, $H$   & S&                      &  59. & $A$, $T_a$, $I$   & L&                     &  94. & $M_a$, $G$, $T_a$   & S&                      &    129.&  $H_a$, $H$, $T_a$   &  L&                      \\ \hline
25. &  $A$, $M_a$, $T_a$ & S&                      &  60. & $A$, $T_b$, $T_c$ & S&                     &  95. & $M_a$, $G$, $T_b$   & U& \cite{WernickUpdate} &    130.&  $H_a$, $H$, $T_b$   &  U& \cite{WernickUpdate} \\ \hline
26. &  $A$, $M_a$, $T_b$ & U& \cite{WernickUpdate} &  61. & $A$, $T_b$, $I$   & S&                     &  96. & $M_a$, $G$, $I$     & S& \cite{WernickUpdate} &    131.&  $H_a$, $H$, $I$     &  S& \cite{WernickUpdate} \\ \hline
27. &  $A$, $M_a$, $I$   & S& \cite{WernickUpdate} &  62. & $O$, $M_a$, $M_b$ & S&                     &  97. & $M_a$, $H_a$, $H_b$ & S&                      &    132.&  $H_a$, $T_a$, $T_b$ &   &                      \\ \hline
28. &  $A$, $M_b$, $M_c$ & S&                      &  63. & $O$, $M_a$, $G$   & S&                     &  98. & $M_a$, $H_a$, $H$   & L&                      &    133.&  $H_a$, $T_a$, $I$   &  S&                      \\ \hline
29. & $A$, $M_b$, $G$    & S&                      &  64. & $O$, $M_a$, $H_a$ & L&                     &  99. & $M_a$, $H_a$, $T_a$ & L&                      &    134.&  $H_a$, $T_b$, $T_c$ &   &                      \\ \hline
30. & $A$, $M_b$, $H_a$  & L&                      &  65. & $O$, $M_a$, $H_b$ & S&                     &  100.& $M_a$, $H_a$, $T_b$ & U& \cite{WernickUpdate} &    135.&  $H_a$, $T_b$, $I$   &   &                      \\ \hline
31. & $A$, $M_b$, $H_b$  & L&                      &  66. & $O$, $M_a$, $H$   & S&                     &  101.& $M_a$, $H_a$, $I$   & S&                      &    136.&  $H$, $T_a$, $T_b$   &   &                      \\ \hline
32. & $A$, $M_b$, $H_c$  & L&                      &  67. & $O$, $M_a$, $T_a$ & L&                     &  102.& $M_a$, $H_b$, $H_c$ & L&                      &    137.&  $H$, $T_a$, $I$     &   &                      \\ \hline
33. & $A$, $M_b$, $H$    & S&                      &  68. & $O$, $M_a$, $T_b$ & U& \cite{WernickUpdate}&  103.& $M_a$, $H_b$, $H$   & S&                      &    138.&  $T_a$, $T_b$, $T_c$ &  U&  \cite{Wernick138}   \\ \hline
34. & $A$, $M_b$, $T_a$  & S&                      &  69. & $O$, $M_a$, $I$   & S&                     &  104.& $M_a$, $H_b$, $T_a$ & S&                      &    139.&  $T_a$, $T_b$, $I$   &  S&                      \\ \hline
35. & $A$, $M_b$, $T_b$  & L&                      &  70. & $O$, $G$, $H_a$   & S&                     &  105.& $M_a$, $H_b$, $T_b$ & S&                      &        &                      &   &                      \\ \hline
\end{tabular}
\end{center}
}
\caption{Wernick's problems and their current status}
\label{table:wernick-list}
\end{table}

An extended list, involving four additional points ($E_a$, $E_b$, $E_c$ ---
three Euler points, which are the midpoints between the vertices and the
orthocenter and $N$ --- the center of the nine-point circle) was presented and
partly solved by Connelly \cite{wernick-connelly}. There are also other variations
of Wernick's list, for instance, the list of problems to be solved given three out
of the following 18 elements: sides $a$, $b$, $c$; angles $\alpha$, $\beta$, $\gamma$;
altitudes $h_a$, $h_b$, $h_c$; medians $m_a$, $m_b$, $m_c$; angle bisectors $t_a$,
$t_b$, $t_c$; circumradius $R$; inradius $r$; and semiperimeter $s$.
There are 186 significantly different non-trivial problems, and it was reported that
(using Wernick's notation) 3 belong to the {\bf R} group, 128 belong to the {\bf S} group,
27 belong to the {\bf U} group, while the status of the remaining ones was unknown
\cite{Berzsenyi}. In addition to the above elements, radiuses of external incircles
$r_a$, $r_b$, $r_c$ and the triangle area $S$ can be also considered, leading to
the list of 22 elements and the total of 1540 triples. Lopes presented solutions to
371 non-symmetric problems of this type \cite{Lopes} and Fursenko considered the list of
(both solvable and unsolvable) 350 problems of this type \cite{Fursenko1,Fursenko2}.

\section{Underlying Geometry Knowledge}
\label{sec:Underlying}

Consider again the problem from Section \ref{sec:introduction} (it is
problem 4 from Wernick's list). One solution is as follows: {\em construct
the midpoint $M_c$ of the segment $AB$ and then construct a point $C$ such that
$\widearrow{M_cG}/\widearrow{M_cC}=1/3$.} Notice that this solution
implicitly or explicitly uses the following:
\begin{enumerate}
\item $M_c$ is the side midpoint of $AB$ (definition of $M_c$);
\item $G$ is the barycenter of $ABC$ (definition of $G$);
\item it holds that $\widearrow{M_cG}/\widearrow{M_cC}=1/3$ (lemma);
\item it is possible to construct the midpoint of a line segment;
\item given points $X$ and $Y$, it is possible to construct a point $Z$,
such that $\widearrow{XY}/\widearrow{XZ}=1/3$.
\end{enumerate}

However, the nature of the above properties is typically not stressed within
solutions of construction problems and the distinctions are assumed. Of course,
given a proper proof that a construction meets the specification, this does not
really affect the quality of the construction, but influences the meta-level
understanding of the domain and solving techniques that it admits. Following
our analyses of Wernick's list, we insist on a clear separation of concepts in
the process of solving construction problems: separation into definitions,
lemmas (geometry properties), and construction primitives. This separation will
be also critical for automating the solving process.

\begin{figure}[h!]
\vspace*{-8mm}
\begin{center}
\input{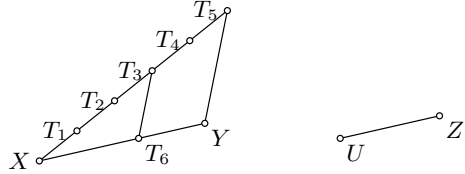}
\end{center}
\vspace*{-6mm}
\caption{Illustration for the construction: given points $X$, $Y$ and $U$,
and a rational number $r$, it is possible to construct a point $Z$ such that:
$\widearrow{XY}/\widearrow{UZ} = r$ (for $r=5/3$)
}
\label{fig:r-construction}
\end{figure}

Our analyses of available solutions of Wernick's problems\footnote{
In addition to 59 solutions available from the Internet \cite{Specht-web}, we
solved 6 problems, which leaves us with 7 solvable problems with no solutions.}
led to the list of 67 high-level construction rules, many of which were based on
complex geometry properties and complex compound constructions. We implemented
a simple forward chaining algorithms using these rules and it was able to solve each
of solvable problems within 1s. Hence, the search over this list of rules is not
problematic --- what is problematic is how to represent the underlying geometry
knowledge and derive this list. Hence, our next goal was to derive high-level
construction steps from the (small) set of definitions, lemmas and construction primitives.
For instance, from the following:
\begin{itemize}
\item it holds that $\widearrow{M_cG} = 1/3\widearrow{M_cC}$ (lemma);
\item given points $X$, $Y$ and $U$, and a rational number $r$, it is
possible to construct a point $Z$ such that: $\widearrow{XY}/\widearrow{UZ} = r$
(construction primitive; Figure \ref{fig:r-construction});
\end{itemize}
we should be able to derive:
\begin{itemize}
\item given $M_c$ and $G$, it is possible to construct $C$.
\end{itemize}

\vspace*{-4mm}

After a careful study, we came to a relatively small list of geometry properties
and primitive constructions needed. In the following, we list all definitions, geometry
properties, and primitive constructions needed for solving most of the problems from
Wernick's list that are currently known to be solvable.
The following notation will be used: $XY$ denotes the line passing through the
distinct points $X$ and $Y$; $\overline{XY}$ denotes the segment with endpoints $X$ and $Y$;
$\widearrow{XY}$ denotes the vector with endpoints $X$ and $Y$; $k(X,Y)$
denotes the circle with center $X$ that passes through $Y$; ${\cal H}(X,Y;Z,U)$
denotes that the pair of points $X$, $Y$ is harmonically conjugate with the pair
of points $Z$, $U$ (i.e., $\widearrow{XU}/\widearrow{UY}=-\widearrow{XZ}/\widearrow{ZY}$);
$s_p(X)$ denotes the image of $X$ in line reflection with respect to a line $p$;
$homothety_{X,r}(Y)$ denotes the image of $Y$ in homothety with respect to a
point $X$ and a coefficient $r$.

\subsection{Definitions}
\label{subsec:defs}

Before listing the definitions used, we stress that we find the standard definition
of the barycenter ({\em the barycenter of a triangle is the intersection of
the medians}) and the like --- inadequate. Namely, this sort of definitions hides
a non-trivial property that all three medians (the lines joining each vertex with
the midpoint of the opposite side) do intersect in one point. Our, constructive
version of the definition of the barycenter says that the barycenter $G$ of a
triangle $ABC$ is the intersection of two medians $AM_a$ and $BM_b$ (it follows
directly from Pasch's axiom that this intersection exists). Several of the
definitions given below are formulated in this spirit. Notice that, following
this approach, in contrast to the Wernick's criterion, for instance, the problems
$(A, B, G)$ and $(A, C, G)$ are not symmetrical (but we do not revise
Wernick's list).

\begin{figure}
\begin{center}
\input{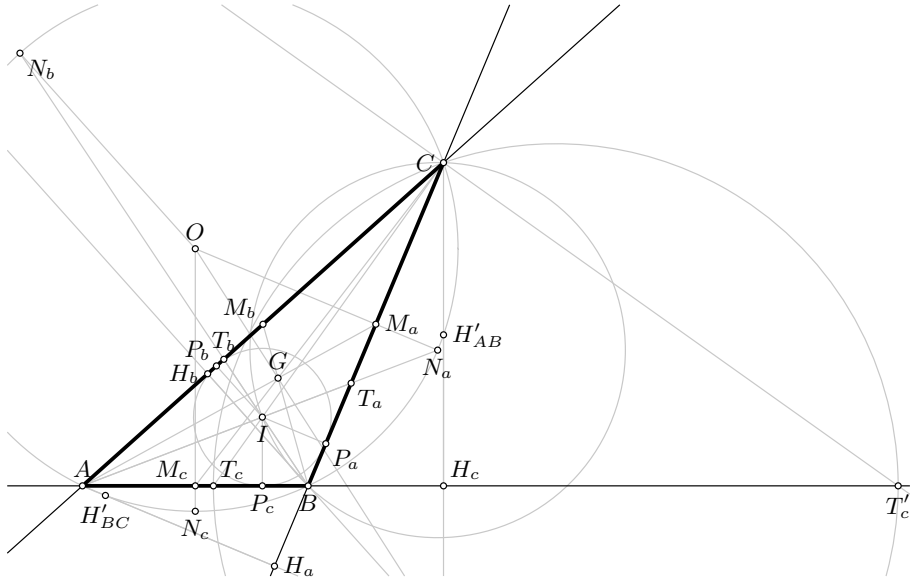}
\end{center}
\vspace*{-6mm}
\caption{Points used in solutions to Wernick's problems}
\label{fig:definitions}
\end{figure}

For a triangle $ABC$ we denote by (along Wernick's notation; Figure
\ref{fig:definitions}):

\begin{enumerate}
\item $M_a$, $M_b$, $M_c$ (the side midpoints): points such that
$\widearrow{BM_a}/\widearrow{BC}=1/2$,
$\widearrow{CM_b}/\widearrow{CA}=1/2$,
$\widearrow{AM_c}/\widearrow{AB}=1/2$;
\item $O$ (circumcenter): the intersection of lines perpendicular at $M_a$ and
$M_b$ on $BC$ and $AC$;
\item $G$ (barycenter): the intersection of $AM_a$ and $BM_b$;
\item $H_a$, $H_b$, $H_c$: intersections of the side perpendiculars with the opposite sides;
\item $H$ (orthocenter): the intersection of $AH_a$ and $BH_b$;
\item $T_a$, $T_b$, $T_c$: intersections of the internal angles bisectors with the opposite sides;
\item $I$ (incenter): the intersection of $AT_a$ and $BT_b$;
\item $T'_a$, $T'_b$, $T'_c$: intersections of the external angles bisectors with the opposite sides;
\item $H_{BC}'$, $H_{AC}'$, $H'_{AB}$: images of $H$ in reflections over lines $BC$, $AC$ and $AB$;
\item $P_a$, $P_b$, $P_c$: feet from $I$ on $BC$, $AC$ and $AB$;
\item $N_a$, $N_b$, $N_c$: intersections of $OM_a$ and $AT_a$, $OM_b$ and $BT_b$, $OM_c$ and $CT_c$.
\end{enumerate}
\vspace*{-7mm}

\vspace*{-6mm}

\subsection{Lemmas}

For a triangle $ABC$ it holds that (Figure \ref{fig:definitions}):

\begin{enumerate}
\item $O$ is on the line perpendicular at $M_c$ on $AB$;
\item $G$ is on $CM_c$;
\item $H$ is on $CH_c$;
\item $I$ is on $CT_c$;
\item $B$ and $C$ are on $k(O,A)$;
\item $P_b$ and $P_c$ are on $k(I,P_a)$;
\item $\widearrow{AG}/\widearrow{AM_a}= 2/3$, $\widearrow{BG}/\widearrow{BM_b}= 2/3$, $\widearrow{CG}/\widearrow{CM_c}= 2/3$;
\item $\widearrow{M_bM_a}/\widearrow{AB}= 1/2$, $\widearrow{M_cM_b}/\widearrow{BC}= 1/2$, $\widearrow{M_cM_a}/\widearrow{AC}= 1/2$;
\item $\widearrow{HG}/\widearrow{HO}= 2/3$;
\item $\widearrow{M_aO}/\widearrow{HA}= 1/2$, $\widearrow{M_bO}/\widearrow{HB}= 1/2$, $\widearrow{M_cO}/\widearrow{HC}= 1/2$;
\item $AB$, $BC$, $CA$ touch $k(I,P_a)$;
\item $N_a$, $N_b$, $N_c$ are on $k(O,A)$;
\item $H'_{BC}, H'_{AC}$, $H'_{AB}$ are on $k(O,A$);
\item $C$, $H_b$, $H_c$ are on $k(M_a,B)$;
      $A$, $H_a$, $H_c$ are on $k(M_b,C)$;
      $B$, $H_a$, $H_b$ are on $k(M_c,A)$;
\item $B,I$ are on $k(N_a,C)$; $C,I$ are on $k(N_b,A)$; $A,I$ are on $k(N_c,B)$;
\item $AH,BH,CH$ are internal angles bisectors of the triangle $H_aH_bH_c$;
\item ${\cal H}(B,C;T_a,T'_a)$, ${\cal H}(A,C;T_b,T'_b)$, ${\cal H}(A,B;T_c,T'_c)$;
\item $A$ is on the circle with diameter $T_aT'_a$;
      $B$ is on the circle with diameter $T_bT'_b$;
      $C$ is on the circle with diameter $T_cT'_c$;
\item $\angle T_cIT_b = \angle BAC/2+\pi/2$; $\angle T_bIT_a = \angle ACB/2+\pi/2$; $\angle T_aIT_c = \angle CBA/2+\pi/2$;
\item The center of a circle is on the side bisector of its arbitrary arc;
\item If the points $X$ and $Y$ are on a line $p$, so is their midpoint;
\item If $\widearrow{XY}/\widearrow{ZW}= r$ then $\widearrow{YX}/\widearrow{WZ}= r$;
\item If $\widearrow{XY}/\widearrow{ZW}= r$ then $\widearrow{ZW}/\widearrow{XY}= 1/r$;
\item If $\widearrow{XY}/\widearrow{ZW}= r$ then $\widearrow{WZ}/\widearrow{YX}= 1/r$;
\item If $\widearrow{XY}/\widearrow{XW}= r$ then $\widearrow{WY}/\widearrow{WX}= 1-r$;
\item If ${\cal H}(X,Y;Z,W)$ then ${\cal H}(Y,X;W,Z)$;
\item If ${\cal H}(X,Y;Z,W)$ then ${\cal H}(Z,W;X,Y)$;
\item If ${\cal H}(X,Y;Z,W)$ then ${\cal H}(W,Z;Y,X)$;
\item If $\widearrow{XY}/\widearrow{XZ}= r$, $Z$ is on $p$, and $Y$ is not on $p$, then
$X$ is on $homothety_{Y,r/(1-r)}(p)$.
\end{enumerate}

All listed lemmas are relatively simple and are often taught in primary or
secondary schools within first lectures on ``triangle geometry''. They can be
proved using a Hilbert's style geometry axioms or by algebraic theorem provers.

\subsection{Primitive Constructions}

We consider the following primitive construction steps:

\begin{enumerate}
\item Given distinct points $X$ and $Y$ it is possible to construct the line $XY$;
\item Given distinct points $X$ and $Y$ it is possible to construct $k(X,Y)$;
\item Given two distinct lines/a line and a circle/two distinct circles that intersect
it is possible to construct their common point(s);
\item Given distinct points $X$ and $Y$ it is possible to construct the side bisector of $\overline{XY}$;
\item Given a point $X$ and a line $p$ it is possible to construct the line $q$ that passes through
$X$ and is perpendicular to $p$;
\item Given distinct points $X$ and $Y$ it is possible to construct the circle with diameter $\overline{XY}$;
\item Given three distinct points it is possible to construct the circle that contains them all;
\item Given points $X$ and $Y$ and an angle $\alpha$ it is possible to construct the set
of (all) points $S$ such that $\angle XSY = \alpha$;
\item Given a point $X$ and a line $p$ it is possible to construct the point $s_p(X)$;
\item Given a line $p$ and point $X$ that does not lie on $p$ it is possible to construct
the circle $k$ with center $X$ that touches $p$;
\item Given a point $X$ outside a circle $k$ it is possible to construct the line $p$ that passes
through $X$ and touches $k$;
\item Given two half-lines with the common initial point, it is possible
to construct an angle congruent to the angle they constitute;
\item Given two intersecting lines it is possible to construct the bisector of internal
angle they constitute;
\item Given one side of an angle and its internal angle bisector it is possible to construct
the other side of the angle;
\item Given a point $X$, a line $p$ and a rational number $r$, it is possible to construct
the line $homothety_{X,r}(p)$;
\item Given points $X$, $Y$, and $Z$, and a rational number $r$ it is possible
to construct the point $U$ such that $\widearrow{UX}/\widearrow{YZ}=r$.
\end{enumerate}

All of the above construction steps can be (most of them trivially) expressed in
terms of straightedge and compass operations. Still, for practical reasons, we use
the above set instead of elementary straightedge and compass operations. These
practical reasons are both more efficient search and simpler, high-level and more
intuitive solutions.

\section{Search Algorithm}
\label{sec:algorithm}

Before the solving process, the preprocessing phase is performed on
the above list of definitions and lemmas. For a fixed triangle $ABC$
all points defined in Section \ref{subsec:defs} are uniquely determined
(i.e., all definitions are instantiated). We distinguish between two types of lemmas:
\begin{description}
\item[instantiated lemmas:] lemmas that describe properties of one or a couple of fixed
objects (lemmas 1-20).
\item[generic lemmas:] lemmas given in an implication form (lemmas 21-29).
These lemmas are given in a generic form and they are instantiated in a preprocessing phase
by all objects satisfying their preconditions. So the instantiations are restricted
with respect to the facts appearing in the definitions or lemmas.
\end{description}

Primitive constructions are given in a generic, non-instantiated form and they get
instantiated while seeking for a construction sequence in the following manner: if
there is an instantiation such that all objects from the premises of the
primitive construction are already constructed (or given by a specification of
the problem) then the instantiated object from the conclusion is constructed, if
not already constructed. However, only a restricted set of objects is constructed
-- the objects appearing in some of the definitions or lemmas. For example,
let us consider the primitive construction stating that for two given points it
is possible to construct the bisector of the segment they constitute. If there would be
no restrictions, the segment bisector would be constructed for each two constructed
points, while many of them would not be used anywhere further. In contrast, this
rule would be applied only to a segment for which its bisector occurs in some
of the definitions or lemmas (for instance, when the endpoints of the segment
belong to a circle, so the segment bisector gives a line to which the center of
the circle belongs to). This can reduce search time significantly, as well as a
length of generated constructions.

The goal of the search procedure is to reach all points required by the
input problem (for instance, for all Wernick's problems, the goal is the same:
construct a triangle $ABC$, i.e., the points $A$, $B$ and $C$).
The search procedure is iterative -- in each step it tries to apply a primitive
construction to the known objects (given by the problem specification or already
constructed) and if it succeeds, the search restarts from the first primitive
construction, in the waterfall manner.
If all required points are constructed, the search stops. If no
primitive construction can be applied, the procedure stops with a failure.
The efficiency of solving, and also the found solution may depend on the order
in which the primitive constructions are listed.

We implemented the above procedure in Prolog.\footnote{The source code is available at
\url{http://argo.matf.bg.ac.rs/?content=downloads}.} At this point the program can solve
58 Wernick's problems, each in less than 1s (for other solvable problems it needs some
additional lemmas).\footnote{Currently, the program cannot solve the following
solvable problems: 27, 43, 55, 57, 58, 69, 76, 82, 87, 96, 101, 126, 131, 139.}
Of course, even with the above restricted search there are redundant construction
steps performed during the search process and once the construction is completed,
all these unnecessary steps are eliminated from the final, ``clean'' construction.
The longest final construction consists of 11 primitive construction steps. Most of
these ``clean'' constructions are the same as the ones that can be found in the
literature. However, for problems with several different solutions, the one found
by the system depends on the order of available primitive constructions/definitions/lemmas
(one such example is given in Section \ref{sec:output}). Along with the construction
sequence, the set of non-degeneracy conditions (conditions that ensure that the
constructed objects do exist, associated with some of construction primitives)
is maintained.

\subsection{Output}
\label{sec:output}

Once a required construction is found and simplified, it can be exported to
different formats. Currently, export to simple natural language form is
supported. For example, the construction for problem 7 $(A,B,H)$ is represented
as follows:

\begin{enumerate}
\item Using the point $A$ and the point $H$, construct the line $AH_a$;
\item Using the point $B$ and the point $H$, construct the line $BH_b$;
\item Using the point $A$ and the line $BH_b$, construct the line $AC$;
\item Using the point $B$ and the line $AH_a$, construct the line $BC$;
\item Using the line $AC$ and the line $BC$, construct the point $C$.
\end{enumerate}

\newpage

\begin{figure}
\begin{minipage}[t]{55mm}
{\small
\begin{verbatim}
% free points
point A 5 5
point B 45 5
point H 32 18
% synthesized construction
line h_a A H
line h_b B H
perp b A h_b
perp a B h_a
intersec C a b
% drawing the triangle ABC
cmark_b A
cmark_b B
cmark_r C
cmark_b H
drawsegment A B
drawsegment A C
drawsegment B C
drawdashline h_a
drawdashline h_b
\end{verbatim}
}
\end{minipage}
\begin{minipage}[t]{52mm}
\vspace{3cm}
\input{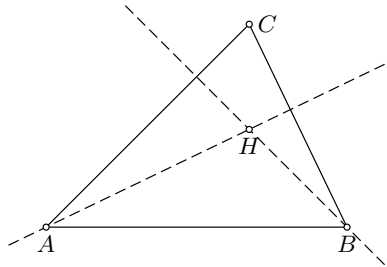}
\end{minipage}
\caption{A GCLC representation (left) and the corresponding illustration (right) for
the solution to Wernick's problem 7}
\label{fig:problem7}
\end{figure}

The generated construction can be (this is subject of our current work) also
represented and illustrated (Figure \ref{fig:problem7}) using the geometry
language GCLC \cite{gclc-jar}.
\vspace*{-4mm}

\vspace*{-4mm}

The above automatically generated solution is also example of a different (and
simpler) solution from the one that can be found on the Internet \cite{Specht-web} and
that we used in building of our system (slightly reformulated in the way to use
our set of primitive constructions):
\begin{enumerate}
\item Using the point $A$ and the point $B$, construct the line $AB$;
\item Using the point $H$ and the line $AB$, construct the line $CH_c$;
\item Using the line $AB$ and the line $CH_c$, construct the point $H_c$;
\item Using the point $H$ and the line $AB$, construct the point $H'_{AB}$;
\item Using the point $A$, the point $B$ and point $H'_{AB}$, construct the circle $k(O,A)$;
\item Using the circle $k(O,A)$ and the line $CH_c$, construct the point $C$.
\end{enumerate}

\subsection{Proving Constructions Correct}

Generated constructions can be proved correct using provers available within
the GCLC tool (the tool provide support for three methods for automated theorem
proving in geometry: the area method, Wu's method, and the Gr\"obner bases method)
\cite{gclc,gclc-ijcar}. For instance, the construction given in Section \ref{sec:output}
in GCLC terms, can be verified using the following additional GCLC code
(note that the given coordinates of the points $A$, $B$ and $H$ are used only
for generating an illustration and are not used in the proving process):

{\small
\begin{verbatim}
% definition of the orthocenter
line _a B C
perp _h_a A _a
line _b A C
perp _h_b B _b
intersec _H _h_a _h_b
% verification
prove { identical H _H }
% alternatively
% prove { perpendicular A H B C }
% prove { perpendicular B H A C }
\end{verbatim}
}

The conjecture that $H$ is indeed the orthocenter of $ABC$ was proved
by Wu's method in 0.01s. Instead of proving that $H$ is identical to the orthocenter,
one could prove that it meets all conditions from the definition of the
orthocenter (which can be more suitable, in terms of efficiency, for automated theorem
provers). For example, the area method proves such two conditions in 0.04s. It also
returns non-degeneracy conditions \cite{area-jar} (needed in the discussion phase):
$A$, $B$ and $H$ are not collinear, neither of the angles $BAH$, $ABH$ is right
angle (additional conditions, such as the condition that the lines $a$ and $b$
from the GCLC construction intersect, are consequences of these conditions).
If $A$ and $B$ are distinct, and $A$ and $H$ are identical, then any point $C$ on
the line passing through $A$ and perpendicular to $AB$ makes a solution, and
similar holds if $B$ and $H$ are identical. Otherwise, if $A$, $B$, and $H$ are
pairwise distinct and collinear or one of the angles $BAH$ and $ABH$ is right
angle, there are no solutions.

\subsection{Re-evaluation}
\label{sec:evaluation}

The presented approach focuses on one sort of triangle construction problems,
but it can be used for other sorts of problems. We re-evaluated our approach
on another corpus of triangle construction problems (discussed in Section
\ref{sec:wernickproblems}). In each problem from this corpus, a task is to
construct a triangle given three lengths from the following set of 9 lengths
of characteristic segments in the triangle:
\begin{enumerate}
\item $|AB|$, $|BC|$, $|AC|$: lengths of the sides;
\item $|AM_a|$, $|BM_b|$, $|CM_c|$: lengths of the medians;
\item $|AH_a|$, $|BH_b|$, $|CH_c|$: lengths of the altitudes.
\end{enumerate}

There are 20 significantly different problems in this corpus and they are all
solvable. This family of problems is substantially different from Wernick's
problems: in Wernick's problems, the task is to construct a triangle based on
the given, located points, while in these problems, the task is to construct a
triangle with some quantities equal to the given ones (hence, the two solutions
to the problem are considered identical if the obtained triangles are congruent).
However, it turns out that the underlying geometry knowledge is mostly shared
\cite{wernick-connelly,Berzsenyi,Lopes,Fursenko1,Fursenko2}. We succeeded to
solve 17 problems from this family, using the system described above and
additional 9 defined points, 2 lemmas, and 8 primitive constructions.
Extensions of the above list of primitive constructions was expected
because of introduction of segment measures.
Since a search space was expanded by adding this new portion of knowledge,
search times increased (the average solving time was 10s) and non-simplified
constructions were typically longer than for the first corpus.
However, simplified constructions are comparable in size with the ones from
the first corpus and also readable.

\section{Future Work}
\label{sec:future}

We plan to work on other corpora of triangle construction problems as
well.\footnote{In our preliminary experiments, our system solved all triangle
construction problems (5 out of 25) in the corpus considered by Gulwani et.al
\cite{GulwaniKT11}; our system can currently solve only a fraction of
135 problems considered by Gao and Chou \cite{GaoC98}, since most of
them are not triangle construction problems or involve the knowledge
still not supported by our system.}
In order to control the search space, the solving system should first detect
the family to which the problem belongs and use only the corresponding rules.
Apart from detecting needed high-level lemmas and rules, we will
try to more deeply explore these lemmas and rules and derive them
from (suitable) axioms and from elementary straightedge and compass
construction steps.

The presented system synthesizes a construction and can use an external
automated theorem prover to prove that the construction meets the
specification (as described in Section \ref{sec:algorithm}; full
automation of linking the solver with automated theorem provers is
subject of our current work). However, the
provers prove only statements of the form: ``if the conditions $\Gamma$ are
met, then the specification is met as well''. They cannot, in a general case,
check if the construction, the conditions $\Gamma$, are consistent (i.e.,
if the points that are constructed do exist). For instance, some
provers cannot check if an intersection of two circles always exist.
We are planning to use proof assistants and our automated theorem
prover for coherent logic \cite{argoclp} for proving that the constructed
points indeed exist (under generated non-degenerate conditions).
With the verified theorem prover based on the area method \cite{area-jar}
or with (more efficient) algebraic theorem proving verified by
certificates \cite{GATPformalization}, this would lead to completely
machine verifiable solutions of triangle construction problems.

\section{Conclusions}
\label{sec:conclusions}

In our work we set up rather concrete tasks: $(i)$ detect geometry knowledge
needed for solving one of the most studied problems in mathematical education
--- triangle construction problems; $(ii)$ develop a practical system for
solving most of these problems. To our knowledge, this is the first systematic
approach to deal with one family of problems (more focused than general
construction problems) and to systemize underlying geometric knowledge.
Our current results lead to a relatively small set of needed definitions,
lemmas, and suitable primitive constructions and to a simple solving
procedure. Generated constructions can be verified using external
automated theorem provers. We believe that any practical solver would need
to treat this detected geometry knowledge one way or another (but trading
off with efficiency). We expect that limited additions to the the geometry
knowledge presented here would enable solving most of triangle construction
problems appearing in the literature.

\paragraph{Acknowledgments.} We are grateful to prof.~Pascal Schreck and
to prof.~Xiao-Shan Gao for providing us lists of construction problems
solved by their systems and for useful feedback.


\begin{thebibliography}{10}

\bibitem{Adler1906}
August Adler.
\newblock {\em Theorie der geometrischen konstruktionen}.
\newblock G\"oschen, 1906.

\bibitem{wernick43}
J.~Anglesio and V.~Schindler.
\newblock Solution to problem 10719.
\newblock {\em American Mathematical Monthly}, 107:952–954, 2000.

\bibitem{Beeson2010}
Michael Beeson.
\newblock Constructive geometry.
\newblock In {\em Proceedings of the Tenth Asian Logic Colloquium}. World
  Scientific, 2010.

\bibitem{Berzsenyi}
George Berzsenyi.
\newblock Constructing triangles from three given parts.
\newblock {\em Quantum}, July/August:396, 1994.

\bibitem{wernick-connelly}
Harold Connelly.
\newblock An extension of triangle constructions from located points.
\newblock {\em Forum Geometricorum}, 9:109–112, 2009.

\bibitem{wernick58}
Harold Connelly, Nikolaos Dergiades, and Jean-Pierre Ehrmann.
\newblock Construction of triangle from a vertex and the feet of two angle
  bisectors.
\newblock {\em Forum Geometricorum}, 7:103–106, 2007.

\bibitem{wernick43a}
Eric Danneels.
\newblock A simple construction of a triangle from its centroid, incenter, and
  a vertex.
\newblock {\em Forum Geometricorum}, 5:53–56, 2005.

\bibitem{DavisTriangleGeometry}
Philip~J. Davis.
\newblock The rise, fall, and possible transfiguration of triangle geometry: A
  mini-history.
\newblock {\em The American Mathematical Monthly}, 102(3):204--214, 1995.

\bibitem{deTemple}
Duane~W. DeTemple.
\newblock Carlyle circles and the lemoine simplicity of polygon constructions.
\newblock {\em The American Mathematical Monthly}, 98(2):97--108, 1991.

\bibitem{constructions-teamat}
Mirjana Djori\'c and Predrag Jani\v{c}i\'c.
\newblock {Constructions, instructions, interactions }.
\newblock {\em {Teaching Mathematics and its Applications}}, 23(2):69--88,
  2004.

\bibitem{Fursenko1}
V.~B. Fursenko.
\newblock Lexicographic account of triangle construction problems (part i).
\newblock {\em Mathematics in schools}, 5:4--30, 1937.

\bibitem{Fursenko2}
V.~B. Fursenko.
\newblock Lexicographic account of triangle construction problems (part ii).
\newblock {\em Mathematics in schools}, 6:21--45, 1937.

\bibitem{GaoC98}
Xiao-Shan Gao and Shang-Ching Chou.
\newblock Solving geometric constraint systems. I. A global propagation
  approach.
\newblock {\em Computer-Aided Design}, 30(1):47--54, 1998.

\bibitem{GaoC98a}
Xiao-Shan Gao and Shang-Ching Chou.
\newblock Solving geometric constraint systems. II. A symbolic approach and
  decision of Rc-constructibility.
\newblock {\em Computer-Aided Design}, 30(2):115--122, 1998.

\bibitem{Grima}
Maria Grima, Gordon J. Pace.
\newblock An Embedded Geometrical Language in Haskell: Construction, Visualisation, Proof.
\newblock {\em Proceedings of Computer Science Annual Workshop}, 2007.

\bibitem{GulwaniKT11}
Sumit Gulwani, Vijay~Anand Korthikanti, and Ashish Tiwari.
\newblock Synthesizing geometry constructions.
\newblock In {\em Programming Language Design and Implementation, PLDI 2011},
  pages 50--61. ACM, 2011.

\bibitem{Guoting}
Chen Guoting.
\newblock Les Constructions G\'eom\'etriques \'a la R\'egle et au Compas par une M\'ethode Alg\'ebrique.
\newblock Master thesis, University of Strasbourg, 1992.

\bibitem{holland}
Gerhard Holland.
\newblock {Computerunterst\"{u}tzung beim L\"{o}sen geometrischer
  Konstruktionsaufgaben}.
\newblock {\em ZDM Zentralblatt f\"{u}r Didaktik der Mathematik}, 24(4),
  1992.

\bibitem{gclc}
Predrag Jani{\v c}i{\'c}.
\newblock {GCLC -- A Tool for Constructive Euclidean Geometry and More than
  That}.
\newblock {\em Proceedings of International Congress of Mathematical Software (ICMS
  2006)}, {\em LNCS} 4151. Springer, 2006.

\bibitem{gclc-ijcar}
Predrag Jani{\v c}i{\'c} and Pedro Quaresma.
\newblock {System description: GCLCprover + GeoThms}.
\newblock {\em International Joint Conference on Automated Reasoning (IJCAR-2006)},
  {\em LNCS} 4130. Springer, 2006.

\bibitem{gclc-jar}
Predrag Jani\v{c}i\'c.
\newblock {Geometry Constructions Language}.
\newblock {\em Journal of Automated Reasoning}, 44(1-2):3--24, 2010.

\bibitem{area-jar}
Predrag Jani\v{c}i\'c, Julien Narboux, and Pedro Quaresma.
\newblock The area method: a recapitulation.
\newblock {\em Journal of Automated Reasoning}, 48(4):489--532, 2012.

\bibitem{Lebesgue}
Henri-L\'eon Lebesgue.
\newblock Le\c{c}ons sur les constructions g\'eom\'etriques.
\newblock Gauthier-Villars, 1950.

\bibitem{Lopes}
L.~Lopes.
\newblock {\em Manuel de Construction de Triangles}.
\newblock QED Texte, 1996.

\bibitem{GATPformalization}
Filip Mari\'c, Ivan Petrovi\'c, Danijela Petrovi\'c, and Predrag Jani\v{c}i\'c.
\newblock Formalization and implementation of algebraic methods in geometry.
\newblock {\em Electronic Proceedings in Theoretical Computer Science}, 79, 2012.

\bibitem{Martin}
George~E. Martin.
\newblock {\em Geometric Constructions}.
\newblock Springer, 1998.

\bibitem{WernickUpdate}
Leroy~F. Meyers.
\newblock Update on William Wernick's ``triangle constructions with three
  located points''.
\newblock {\em Mathematics Magazine}, 69(1):46--49, 1996.

\bibitem{Pambuccian08}
Victor Pambuccian.
\newblock Axiomatizing geometric constructions.
\newblock {\em Journal of Applied Logic}, 6(1):24--46, 2008.

\bibitem{SchreckPhD}
Pascal Schreck.
\newblock {\em Constructions \`a la r\`egle et au compas}.
\newblock PhD thesis, University of Strasbourg, 1993.

\bibitem{Specht-web}
Eckehard Specht.
\newblock Wernicks liste.
\newblock \url{http://hydra.nat.uni-magdeburg.de/wernick/}.
\newblock In German.

\bibitem{GaloisTheory}
Ian Stewart.
\newblock {\em Galois Theory}.
\newblock Chapman and Hall Ltd., 1973.

\bibitem{argoclp}
Sana Stojanovi\'c, Vesna Pavlovi\'c, and Predrag Jani{\v c}i{\' c}.
\newblock A coherent logic based geometry theorem prover capable of producing
  formal and readable proofs.
\newblock {\em Automated Deduction in Geometry}, {\em LNCS} 6877. Springer, 2011.

\bibitem{Wernick138}
Alexey~V. Ustinov.
\newblock On the construction of a triangle from the feet of its angle
  bisectors.
\newblock {\em Forum Geometricorum}, 9:279–280, 2009.

\bibitem{Wernick}
William Wernick.
\newblock Triangle constructions vith three located points.
\newblock {\em Mathematics Magazine}, 55(4):227--230, 1982.

\bibitem{wernick57}
Paul Yiu.
\newblock Elegant geometric constructions.
\newblock {\em Forum Geometricorum}, 5:75–96, 2005.

\end{thebibliography}

\end{document}